%% file: main.tex
\documentclass{ceurart}

\usepackage{booktabs}
\usepackage{array}
\usepackage{amsmath}
\usepackage{amssymb}
\usepackage{placeins}

\begin{document}

\copyrightyear{2026}
\copyrightclause{Copyright for this paper by its authors.
  Use permitted under Creative Commons License Attribution 4.0
  International (CC BY 4.0).}

\conference{ReNeuIR at SIGIR 2026: The Fifth Workshop on Reaching Efficiency in Neural Information Retrieval, July 24, 2026, Melbourne, Australia}

\title{StaRSE: Compact Russian Sentence Embeddings with a Sign-Coded Static Encoder}

\author[1]{Boris Malashenko}[%
email=btmalashenko@itmo.ru,
]
\author[1]{Ivan Jarsky}[%
email=ivanjarsky@itmo.ru,
]
\author[1]{Egor Soldatov}[%
email=501866@niuitmo.ru,
]
\author[1]{Vladislav Luzhnov}[%
email=501693@niuitmo.ru,
]
\author[1]{Sviatoslav Stumpf}[%
email=sastumpf@itmo.ru,
]
\author[1]{Vladislav Shmatkov}[%
email=vnshmatkov@itmo.ru,
]
\author[1]{Viacheslav Shalamov}[%
email=shalamov@itmo.ru,
]
\author[1]{Andrey Stankevich}[%
email=stankev@itmo.ru,
]
\author[1]{Valeria Efimova}[%
email=vefimova@itmo.ru,
]
\address[1]{ITMO University, Saint Petersburg, Russia}

\begin{abstract}
Transformer sentence encoders provide strong retrieval quality, but their inference cost becomes a bottleneck when embeddings must be produced on CPUs or refreshed frequently.
We introduce StaRSE, a compact Russian static sentence encoder that removes Transformer inference at deployment time: token directions are stored as sign bits, token norms are stored separately, and sentences are encoded by mean pooling over norm-scaled sign directions.
StaRSE is initialized from ruBERT-base, fine-tuned on Russian adjacent-sentence and task-shaped pairs, and post-trained with a contrastive stage that directly optimizes the sign-projected token table.
On RuMTEB, StaRSE reaches a task mean of $51.16$, $7.98$ points above the strongest included static baseline. The packaged model occupies $11.3$ MiB and has a mean CPU latency of $0.08$ ms per sentence, compared with an estimated $55.7$ MiB 16-bit parameter footprint and $1.43$ ms mean CPU latency for the closest-quality contextual encoder.
Ablations examine adjacent-sentence supervision, token norms, and direct sign-projected post-training.
\end{abstract}


\begingroup
\hfuzz=500pt
\maketitle
\endgroup

\input{sections/01_introduction}
\input{sections/02_related_work}
\input{sections/03_method}
\input{sections/04_experiments}

\input{sections/05_analysis}

\input{sections/06_conclusion}

\section*{Limitations}
This work focuses on Russian, a large retrieval language with mature contextual encoders but comparatively few language-specific static sentence encoders.
Applying the recipe to other languages would require separate validation of the tokenizer, filtering thresholds, source mixture, and training settings.
StaRSE is also deliberately static: mean pooling over token embeddings cannot model word order, syntax, or contextual sense disambiguation in the way a Transformer can.

The main comparison and sign-projection ablations use completed full $23$-task RuMTEB runs, but the large contrastive runs are single-seed and no run-to-run noise floor has been measured. The data-construction and distillation blocks use different training budgets, while the continuous initializer and final sign-projected stage use different source mixtures; these comparisons are therefore diagnostic and do not isolate the contribution of one data source. In addition, the final sign-projected hyperparameters were selected with an exploratory Optuna sweep over the full RuMTEB mean, so the main score is not a held-out estimate of generalization.

Our efficiency measurements cover encoder latency and encoder storage rather than a complete retrieval service. The RuMTEB lexical rankers use fixed settings rather than per-task tuning. The reported MLDR fusion setting was selected on the same evaluation pool, so its results are descriptive. Per-task tuning and analysis of the query or domain conditions under which lexical rankers complement StaRSE remain future work.

The compact model evaluated in this paper is a sign-coded token table combined with real-valued mean pooling and $\ell_2$-normalized sentence vectors.
Strict one-bit-per-dimension sentence representations would reduce stored sentence vectors further, but in our configuration they cost $3.84$ RuMTEB points (Table~\ref{tab:sign-projection-ablations}).
Closing this remaining gap likely requires explicitly training the sentence vectors to be signed after pooling or rethinking how token norms participate in pooling; this is left for future work.

\section*{Ethics Statement}
StaRSE is an encoder rather than a generative model: it maps text to vectors and does not produce fluent text or answer user queries by itself.
The main ethical risks therefore come from representation bias and downstream retrieval or ranking use.
The training mixture includes web-derived Russian text, which may contain social biases, offensive language, or low-quality passages.
We use Cultura RU Edu as the broad web-text source and document the supervised and retrieval sources mixed with it.
We do not redistribute the source corpus text.
Users who deploy the encoder in search, recommendation, or classification systems should evaluate the full downstream system for bias, robustness, and privacy-sensitive behavior in its target domain.

\section*{Declaration on Generative AI}
During the preparation of this work, the authors used OpenAI Codex and ChatGPT for language editing, wording suggestions, consistency checks, and LaTeX polishing. After using these tools, the authors reviewed and edited the content as needed and take full responsibility for the publication's content.

\begin{acknowledgments}
This work supported by the Ministry of Economic Development of the Russian
Federation (IGK \mbox{000000C313925P4C0002}), agreement \mbox{No139-15-2025-010}.
\end{acknowledgments}

\bibliography{custom}

\end{document}

%% file: sections/01_introduction.tex
\section{Introduction}

Text embedding models are now a standard component of neural retrieval systems, semantic search, clustering, deduplication, recommendation, and retrieval-augmented generation.
The strongest general-purpose embedding models are contextual encoders, typically trained or adapted with sentence-level contrastive objectives \citep{reimers-gurevych-2019-sentence-bert,gao-etal-2021-simcse,wang-etal-2024-e5,li-etal-2023-gte}.
Their quality, however, comes with a deployment cost: every query and document requires a Transformer forward pass~\citep{vaswani-etal-2017-attention}.
This cost is often acceptable for reranking~\citep{nogueira-cho-2019-passage} or small collections, but it becomes restrictive when embeddings must be produced on CPU, refreshed frequently, served under tight latency budgets, or computed for very large text collections.

Static embedding models, from word2vec \citep{mikolov-etal-2013-distributed}, GloVe \citep{pennington-etal-2014-glove}, and fastText \citep{bojanowski-etal-2017-enriching} to sentence-level and recent static embedding variants \citep{pagliardini-etal-2018-sent2vec,wada-etal-2025-static,minishlab2024model2vec}, offer a different point in this quality-cost trade-off.
At inference time, they use a fixed vocabulary, a simple lookup table, pooling over token vectors, and normalization.
This makes inference cheap and predictable, especially on CPUs and resource-constrained hardware.
The price is a strict architectural bottleneck: a mean-pooled static embedding model cannot model word order, contextual word senses, or compositional interactions in the way a Transformer can.
The practical question is therefore not whether static embedding models should replace contextual encoders, but whether they can provide useful retrieval quality at much lower serving cost.

This question is underexplored for Russian and, more broadly, for language-specific efficient retrieval outside English.
Recent Russian sentence transformers include strong contextual models such as FRIDA \citep{snegirev-etal-2025-russian} and GigaEmbeddings \citep{kolodin-etal-2025-gigaembeddings}.
Russian static embedding models, however, remain mostly word-level, while multilingual static embedding models are not optimized specifically for Russian morphology and vocabulary coverage.
We treat Russian as one instance of a broader setting: languages with large text corpora but limited high-quality labeled sentence pairs, for which a reusable static-encoder training recipe may be useful.

\begin{figure}[t]
\centering
\includegraphics[width=0.88\linewidth]{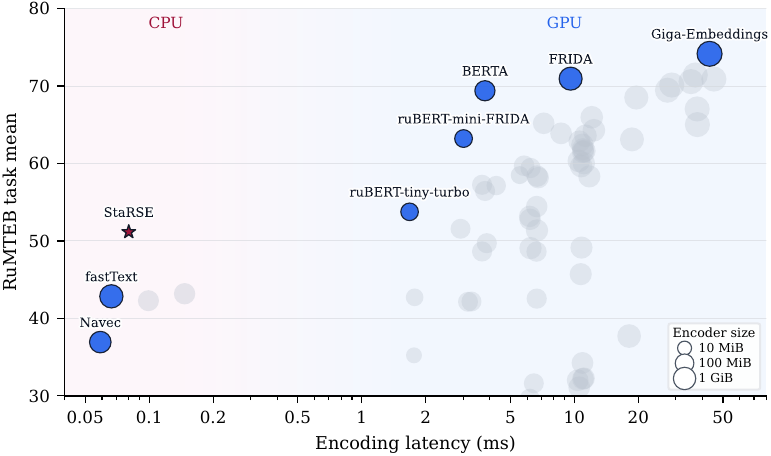}
\caption{
Quality on RuMTEB \citep{snegirev-etal-2025-russian} versus mean end-to-end single-sentence encoding latency in milliseconds (log scale; lower is better).
Static encoders are measured on CPU and contextual encoders on GPU, so the two device groups represent different deployment regimes.
Marker area shows encoder footprint in MiB on a log scale: contextual footprints are estimated as parameter count times two bytes, while static footprints are measured packages.
Labeled blue points form the nondominated frontier of the displayed score--latency measurements under this mixed-device diagnostic.
The star marks StaRSE.
}
\label{fig:rumteb-cpu-latency}
\end{figure}

We introduce \textbf{StaRSE}, a static sentence encoder for Russian with a sign-coded token table: token directions are stored as sign bits, while per-token scalar norms are stored separately. The pipeline deliberately combines established ingredients from static embeddings and contrastive sentence embedding training: Transformer input embeddings, token-level dimensionality reduction, frequency weighting, adjacent-sentence positives, and InfoNCE-style optimization. The main StaRSE-specific component is the compact sign-coded token table with mean pooling over norm-scaled sign directions and a final training stage that computes the contrastive loss through the sign-projected table instead of signing a continuous table only after training. At deployment time, StaRSE stores only the compact token table and encodes text by pooling norm-scaled sign directions, avoiding Transformer inference for both queries and documents.
Figure~\ref{fig:rumteb-cpu-latency} shows this trade-off numerically: StaRSE reaches $51.16$ task mean on RuMTEB, $7.98$ points above the strongest included static encoder, and remains within $2.60$ points of ruBERT-tiny-turbo. In Table~\ref{tab:encoding-efficiency}, StaRSE's measured package is $11.3$ MiB, while the estimated 16-bit parameter footprint of ruBERT-tiny-turbo is $55.7$ MiB. Under the same CPU protocol, StaRSE has about $18\times$ lower mean single-sentence latency ($0.08$ versus $1.43$ ms).
Larger contextual encoders remain substantially more accurate, but they occupy a different cost range.

Our contributions are as follows.
\begin{enumerate}
\item We introduce StaRSE, a compact static sentence encoder for Russian. StaRSE reaches $51.16$ RuMTEB: $7.98$ points above the strongest included static encoder and $2.60$ points below the closest included contextual encoder by RuMTEB score.
\item We provide an empirical recipe for carrying out the full training pipeline for a language-specific static sentence encoder, including tokenizer and initialization choices, document-adjacency mining, task-shaped pair construction, and training-stage decisions.
\item We propose contrastive post-training for static embedding models that directly optimizes the sign-projected token geometry.
\end{enumerate}
The evaluated SentenceTransformers-compatible StaRSE model weights\footnote{\url{https://huggingface.co/BorisTM/starse}} and training code\footnote{\url{https://github.com/BlessedTatonka/StarSE}} are publicly available.

%% file: sections/02_related_work.tex
\section{Related Work}

Static word embeddings were the dominant vector representation before the rise of contextual encoders.
Models such as word2vec \citep{mikolov-etal-2013-distributed}, GloVe \citep{pennington-etal-2014-glove}, and fastText \citep{bojanowski-etal-2017-enriching} assign one vector to each vocabulary unit.
Sentence or document vectors were then commonly built compositionally, for example by averaging word vectors, adding $n$-gram features as in Sent2Vec \citep{pagliardini-etal-2018-sent2vec}, or applying simple weighting and post-processing heuristics.
These approaches are attractive because inference reduces to lookup and pooling, but their tables are usually optimized for word-level distributional semantics rather than for general-purpose sentence embeddings.

Contextual encoders now provide the strongest sentence representations.
BERT \citep{devlin-etal-2019-bert} introduced contextual token representations, and Sentence-BERT \citep{reimers-gurevych-2019-sentence-bert} adapted Transformer encoders for efficient sentence-level similarity.
We use \emph{Sentence Transformer} (ST) as a shorthand for Transformer encoders trained or adapted to produce sentence-level embeddings directly.
Later STs and related contextual embedding systems such as SimCSE \citep{gao-etal-2021-simcse}, E5 \citep{wang-etal-2024-e5}, and GTE \citep{li-etal-2023-gte} improved embedding quality through successive changes to the training pipeline and data construction.
MTEB \citep{muennighoff-etal-2023-mteb} broadened evaluation beyond STS to information retrieval (IR), reranking, classification, clustering, and pair classification, including natural language inference (NLI)-style tasks.
This progress made STs the default choice for high-quality embeddings, while static models remained attractive when CPU latency, memory use, indexing cost, or hardware availability are limiting factors.

Recent work has revisited static embedding models by reusing representations from pretrained Transformer models.
WordLlama recycles the input token embeddings of LLaMA-family models and trains a lightweight context-free text embedding model from them \citep{miller2024wordllama}.
Model2Vec instead distills a static vocabulary table from an existing sentence embedding model, producing a fast lookup-and-pooling model without running the teacher at inference time \citep{minishlab2024model2vec}.
A third line starts from contextualized word representations: contextual models assign different vectors to the same word in different sentences \citep{ethayarajh-2019-contextual}, and these contextual vectors can be reduced to static word representations by aggregating occurrences \citep{bommasani-etal-2020-interpreting}.
Wada and colleagues apply this idea to sentence embeddings by collecting many corpus occurrences of each word, encoding them in context, and averaging the resulting token representations into one static vector \citep{wada-etal-2025-static}.
These recipes differ in what they require.
Input-embedding reuse and teacher distillation need a pretrained model and its vocabulary, whereas occurrence-based extraction also needs a corpus with enough contexts for each vocabulary item.
Such extraction can be computationally expensive because it requires many Transformer forward passes to collect last-hidden-state token representations before the static table can be built.

Dimensionality reduction and token weighting are central to the initialization and post-processing used in these static models.
WordLlama and Model2Vec apply PCA to the token embedding matrix, while \citet{wada-etal-2025-static} argue for sentence-level PCA, in which PCA is fitted to averaged sentence representations so that the retained axes better reflect sentence-level variance.
In both cases, PCA can be combined with All-but-the-Top (ABTT) post-processing, which discards a small number of dominant principal components instead of simply keeping the leading ones \citep{mu-viswanath-2018-abtt}.
This can reduce the influence of common directions that appear across many embeddings and can otherwise dominate cosine similarities.
Frequency-based weighting addresses a different issue: very frequent tokens often contribute less semantic specificity to an averaged sentence vector.
SIF \citep{arora-etal-2017-sif} downweights frequent tokens using corpus frequency.
When reliable counts are unavailable, a Zipfian prior over vocabulary rank provides a simple approximation \citep{piantadosi-2014-zipf}.

After initialization and post-processing, static embedding models are often trained with either distillation or contrastive learning.
Knowledge distillation transfers information from a stronger teacher to a smaller student \citep{hinton-etal-2015-distilling}.
For STs, this can mean matching teacher sentence vectors, matching pairwise teacher similarities within a batch, or aligning multilingual sentence spaces \citep{reimers-gurevych-2020-making,wada-etal-2025-static}.
Distillation is attractive because it can create a static model from a teacher and vocabulary without a labeled pair dataset, but direct vector matching constrains the student to approximate a teacher geometry with much lower capacity.
Contrastive learning instead trains the model to score positive pairs above in-batch or explicitly sampled negatives, as in InfoNCE-style embedding fine-tuning \citep{oord-etal-2018-cpc,sturua-etal-2024-jina-v3}.

Embedding compression is another way to reduce memory and search cost for embedding models.
Some text-retrieval and sentence-representation systems binarize final dense sentence or passage vectors so nearest-neighbor search can use Hamming distance or equivalent bit operations \citep{shen-etal-2019-learning,hamann-etal-2019-hamming,yamada-etal-2021-efficient}.
In our work, StaRSE uses a different form of compression: it stores sign-projected token directions and separate scalar token norms, while the pooled sentence vector remains real-valued.
Vector and weight quantization reduce memory and arithmetic precision, as in quantization-aware training for neural network inference \citep{jacob-etal-2018-quantization}.
Quantization-aware methods are useful when post-hoc quantization would change the representation seen by the loss: training with the quantized forward path lets the objective optimize the representation used at inference.

Russian embedding research follows the broader contextual trend.
RuMTEB provides a Russian version of MTEB with the same broad task families \citep{snegirev-etal-2025-russian}, while RusBEIR collects Russian information-retrieval tasks for retrieval-focused evaluation \citep{kovalev-etal-2025-rusbeir}.
Recent Russian sentence-embedding work includes FRIDA \citep{snegirev-etal-2025-russian} and GigaEmbeddings \citep{kolodin-etal-2025-gigaembeddings}.
GigaEmbeddings combines large-scale contrastive pre-training with hard-negative and multitask fine-tuning, synthetic data, and efficient contextual pooling.
Russian static embedding models are mostly word-level: Navec\footnote{\url{https://github.com/natasha/navec}} and DeepPavlov\footnote{\url{https://docs.deeppavlov.ai/en/master/features/pretrained_vectors.html}} both provide compact Russian word embeddings.
These models are lightweight, but using them for sentence-level semantics typically requires adapting word-level vectors rather than using a model trained for sentence-level similarity.
When static text embedders are multilingual, their vocabulary and capacity are shared across languages rather than targeted to Russian.
This leaves room for a Russian-specific static sentence encoder trained directly for sentence-level retrieval.

%% file: sections/03_method.tex
\section{Method}

StaRSE is built in three steps.
First, we choose a tokenizer vocabulary and initialize the static token table from a pretrained Transformer encoder input embedding matrix (Sec.~\ref{sec:method-init}).
Second, we construct weak positive pairs from source documents by pairing adjacent sentences (Sec.~\ref{sec:method-mining}).
Third, we train the table with a bidirectional InfoNCE loss and then continue training with the sign-projected token table used at inference time (Sec.~\ref{sec:method-training}).

\subsection{Vocabulary Selection and Embedding Initialization}
\label{sec:method-init}

The first design choice is the vocabulary.
A static embedding model inherits both its coverage and its tokenization behavior from a fixed vocabulary $V$.
The tokenizer should keep target language text reasonably compact, while preserving lexical distinctions that matter for retrieval.
We therefore start from a pretrained Transformer and reuse its tokenizer vocabulary $V$ together with its input embedding matrix $Q\in\mathbb{R}^{|V|\times D}$.

Given $V$, StaRSE is a static embedding model with embedding table $E \in \mathbb{R}^{|V|\times d}$.
For a non-empty tokenized text $x=(w_1,\ldots,w_n)$, the unnormalized sentence vector is the mean of the corresponding token vectors:
\begin{equation}
  h(x)=\frac{1}{n}\sum_{i=1}^{n}E_{w_i}.
\end{equation}
The final representation is length-normalized:
\begin{equation}
  z(x)=\frac{h(x)}{\max(\lVert h(x)\rVert_2,\varepsilon)}.
\end{equation}
For two texts $x$ and $y$, we use the dot product of their normalized vectors as the similarity score, $s(x,y)={z(x)}^\top z(y)$.
Because both vectors are length-normalized, this is cosine similarity.

To initialize $E$, we extract $Q$ from the selected Transformer and project it to the target dimensionality with token-level PCA. Instead of keeping the first $d$ principal components, we apply ABTT post-processing \citep{mu-viswanath-2018-abtt}. We discard the first $r$ dominant components and keep the next $d$ components. We then apply SIF-style frequency weighting. For the token at position $i\in\{1,\ldots,|V|\}$ in the tokenizer's vocabulary order, we use the normalized Zipf proxy $p_i=i^{-\beta}/\sum_{j=1}^{|V|}j^{-\beta}$ and rescale its projected row as $E_i\leftarrow aE_i/(a+p_i)$, where $a$ is the SIF smoothing coefficient and $\beta$ is the Zipf exponent.

\subsection{Document Pair Construction}
\label{sec:method-mining}

Curated sentence-pair datasets provide high-quality supervision, but for most languages there are few large, diverse, and reliably labeled pair collections.
This is especially limiting for a general static sentence encoder.
Unlike a Sentence Transformer, StaRSE does not retain a pretrained contextual encoder at inference time.
The input embedding matrix gives a lexical and subword-level starting point, but sentence-level similarity must be induced in the token table itself during training.

Document collections provide a broad weakly supervised signal for this purpose.
The use of adjacent sentences follows the same basic intuition as BERT's next sentence prediction objective \citep{devlin-etal-2019-bert}: consecutive text segments often carry discourse information.
We use this signal differently.
Instead of training a classifier to decide whether two segments are consecutive, we use adjacent sentences as positive pairs for contrastive training.
Adjacent sentences are not guaranteed to be paraphrases: they may continue the same topic or discourse without expressing the same meaning.
Nevertheless, document adjacency preserves topic continuity, entities, discourse relations, and local semantic coherence more often than random cross-document pairing.
Section~\ref{sec:adjacent-pair-signal} examines this assumption empirically with a diagnostic sentence encoder.
We therefore treat adjacent sentences as weak positives rather than exact semantic equivalents.

Given a collection of source documents, we segment each document into sentences and form candidate positive pairs from consecutive sentences.
Each pair is $(u_i,u_{i+1})$, where both sentences come from the same document and preserve their original order.
To remove accidental transitions, sentence-splitting artifacts, and pairs with no shared lexical content, we apply a lightweight lemma-overlap filter.
Let $L(x)$ be the set of normalized non-stopword lemmas in sentence $x$. For a candidate pair $(u,v)$, we compute the Dice overlap
\begin{equation}
  \operatorname{LO}(u,v)
  =
  \frac{2|L(u)\cap L(v)|}{|L(u)|+|L(v)|},
\end{equation}
with zero overlap when the denominator is zero.
For the main Cultura pool, we keep only pairs whose lemma overlap is at least $0.20$.
This filter is deliberately lexical and inexpensive: it is used as a fast corpus-cleaning step before contrastive training, not as a semantic teacher or hard-mining procedure.

\subsection{Contrastive Training}
\label{sec:method-training}

We train the static embedding table on text pairs with a bidirectional InfoNCE objective, following the pairwise embedding fine-tuning setup used in recent embedding models \citep{oord-etal-2018-cpc,sturua-etal-2024-jina-v3}.
Let
$\mathcal{B}=((p_1,q_1),\ldots,(p_k,q_k))$ be a batch of $k$ positive pairs. For a scoring function $s(\cdot,\cdot)$ and one batch direction $\mathcal{B}=((x_1,y_1),\ldots,(x_k,y_k))$, the InfoNCE term is
\begin{equation}
  \mathcal{L}_{\mathrm{NCE}}(\mathcal{B};s)
  =
  -\frac{1}{k}\sum_{i=1}^{k}
  \log
  \frac{\exp(s(x_i,y_i)/\tau)}
       {\sum_{i'=1}^{k}\exp(s(x_i,y_{i'})/\tau)},
\end{equation}
where $\tau$ is the temperature.
For each anchor $x_i$, the paired text $y_i$ is the positive example and the other texts $\{y_{i'}: i'\neq i\}$ in the same batch act as in-batch negatives.
Let $\mathcal{B}^\dagger=((q_1,p_1),\ldots,(q_k,p_k))$ denote the same batch with the order of each pair swapped.
The symmetric pair loss is
\begin{equation}
  \mathcal{L}_{\mathrm{pairs}}(\mathcal{B};s)
  =
  \frac{1}{2}
  \left[
  \mathcal{L}_{\mathrm{NCE}}(\mathcal{B};s)
  + \mathcal{L}_{\mathrm{NCE}}(\mathcal{B}^\dagger;s)
  \right].
\end{equation}
For the continuous StaRSE model, $s(\cdot,\cdot)$ is the cosine similarity defined in Section~\ref{sec:method-init}.
The symmetric loss lets each sentence in a pair serve once as the anchor.
The same objective applies to adjacent-sentence pairs and to curated sentence-pair batches.

\textbf{Sign-projected post-training.} The final compact model is obtained through quantization-aware post-training.
After contrastive fine-tuning, let $E^{(0)}$ denote the learned real-valued table.
The compact StaRSE checkpoint is initialized from $E^{(0)}$.
In this stage the optimizer still updates one real-valued embedding matrix $E\in\mathbb{R}^{|V|\times d}$, but every forward pass computes sentence vectors through sign-projected token directions.
This makes the training loss act on the same compact representation that is stored for inference.

For each row of the current table, we separate its norm from its direction:
\begin{equation}
  E_i = \rho_i u_i,\qquad
  \rho_i = \max(\lVert E_i \rVert_2,\varepsilon),\qquad
  u_i=E_i/\rho_i.
\end{equation}
The norm $\rho_i$ sets how strongly token $i$ contributes to a mean-pooled sentence vector, while the direction $u_i$ is the part that will be sign-projected.
These quantities are not stored as separate parameters: they are recomputed from the current $E_i$ at every step, and ordinary optimizer updates on $E$ can change both the token norm and direction.

The forward path uses the sign-projected token table
\begin{equation}
  E_i^B =
  \rho_i \frac{\operatorname{sign}(u_i)}{\sqrt d}.
\end{equation}
Equivalently, for $b_i=\operatorname{sign}(u_i)/\sqrt d$, the pre-normalization pooled vector is
\begin{equation}
  h_B(x)=\frac{1}{n}\sum_{t=1}^{n}\rho_{w_t} b_{w_t}.
\end{equation}
Thus $\rho_i$ determines how much a token contributes to the pooled vector, while $b_i$ determines the sign direction contributed by that token.
Let $f_B(x)$ be the same mean-pooling and $\ell_2$-normalization encoder as in Section~\ref{sec:method-init}, but using $E^B$.
StaRSE sign-projects token directions, not final sentence vectors: after pooling and length normalization, $f_B(x)$ is still a real-valued vector computed from signs plus one scalar norm per token.
The sign-projected similarity is
\begin{equation}
  s_B(x,y) = {f_B(x)}^\top f_B(y).
\end{equation}
The sign projection is used with a straight-through estimator: the forward pass substitutes $\operatorname{sign}(u_i)/\sqrt d$ for the underlying unit row while the backward pass treats it as the identity on $u_i$, so the gradient with respect to $E_i$ flows through both the direction $u_i$ and the magnitude $\rho_i$ even though the forward path is discrete.

For a batch direction $\mathcal{B}=((x_1,y_1),\ldots,(x_k,y_k))$, let $S_B\in\mathbb{R}^{k\times k}$ be the sign-projected score matrix with ${(S_B)}_{ij}=s_B(x_i,y_j)$, and let $S_T$ be the analogous score matrix from the frozen continuous StaRSE model $E^{(0)}$.
We denote by $P_T$ and $P_B$ the collections of row-wise and column-wise distributions induced by these matrices, using
\begin{equation}
  \sigma_r^i(S;\tau_T)=\operatorname{softmax}(S_{i,:}/\tau_T),
  \qquad
  \sigma_c^j(S;\tau_T)=\operatorname{softmax}(S_{:,j}/\tau_T).
\end{equation}
The teacher distribution term is a row-and-column KL divergence,
\begin{equation}
  \begin{aligned}
  \mathrm{KL}(P_T\|P_B)
  &=
  \frac{1}{2k}\sum_{i=1}^{k}
  \mathrm{KL}\!\left(\sigma_r^i(S_T;\tau_T)\middle\|\sigma_r^i(S_B;\tau_T)\right) \\
  &\quad+
  \frac{1}{2k}\sum_{j=1}^{k}
  \mathrm{KL}\!\left(\sigma_c^j(S_T;\tau_T)\middle\|\sigma_c^j(S_B;\tau_T)\right).
  \end{aligned}
\end{equation}
This term encourages the sign-projected table to match the continuous model's row-wise and column-wise similarity rankings within the batch.

The final StaRSE stage combines two loss terms: the symmetric contrastive pair loss on the sign-projected table and the teacher-distribution term:
\begin{equation}
  \label{eq:starse-loss}
  \mathcal{L}_{\mathrm{StaRSE}}
  =
  \alpha\,\mathcal{L}_{\mathrm{pairs}}(\mathcal{B};s_B)
  + \lambda_T\,\mathrm{KL}(P_T \| P_B).
\end{equation}
Here $\alpha$ and $\lambda_T$ weight the contrastive pair loss and the teacher-distribution loss, respectively.

%% file: sections/04_experiments.tex
\section{Experiments}

This section describes the experimental setup and results for StaRSE.
We specify the training data, model configuration, baselines, and evaluation protocol, then report RuMTEB, long-document retrieval, and encoding-efficiency results.
All RuMTEB scores and nDCG@$10$ values are reported on a $0$--$100$ scale throughout the paper; their differences are expressed in points on this scale.

\subsection{Training Data}
\label{sec:training-data}

The contrastive fine-tuning recipe combines adjacent sentence pairs from Cultura RU Edu\footnote{\url{https://huggingface.co/datasets/deepvk/cultura_ru_edu}} with Russian retrieval and paraphrase-style pair sources.
We use Cultura as the broad text source because adjacent sentences expose entities, topics, and discourse transitions that are less common in small supervised similarity datasets.
Following Section~\ref{sec:method-mining}, we use the Razdel library\footnote{\url{https://github.com/natasha/razdel}} to split Cultura documents into sentences and form candidate pairs from consecutive sentences $(s_i,s_{i+1})$ in the same document. Cultura pairs are retained with the lemma-overlap filter from Section~\ref{sec:method-mining}.

The remaining sources provide more task-shaped positives.
From the Russian subset of the mMARCO IR passage-ranking dataset \citep{bonifacio2021mmarco}, we use the query and the positive passage from each training triple.
From the ru-HNP paraphrase dataset,\footnote{\url{https://huggingface.co/datasets/deepvk/ru-HNP}} we treat each source text and its positive paraphrases as a positive set and form all unordered pairs within that set. From ru-paraphrase-NMT-Leipzig,\footnote{\url{https://huggingface.co/datasets/cointegrated/ru-paraphrase-NMT-Leipzig}} a paraphrase/translation dataset, we use the original and back-translated pair.
Basic normalization removes empty, identical, and overlong pairs. During continuous training, a source-balanced sampler prevents the smaller supervised datasets from being overwhelmed by Cultura.

\subsection{Model and Training Setup}
\label{sec:training-setup}

StaRSE uses the WordPiece tokenizer and vocabulary of ruBERT-base\footnote{\url{https://huggingface.co/ai-forever/ruBert-base}}, with $120{,}138$ tokens.
The final model keeps this vocabulary fixed and learns one table row for each token.
Section~\ref{sec:model-selection} provides the development results behind the tokenizer and initialization choices.

We initialize the final $512$-dimensional table from the corresponding input embedding matrix using token-level PCA, ABTT, and SIF-Zipf weighting, as selected in Section~\ref{sec:model-selection}.

Continuous contrastive training has two phases, both optimizing only the symmetric pair loss $\mathcal{L}_{\mathrm{pairs}}(\mathcal{B};s)$ from Section~\ref{sec:method-training}. The first phase runs for $20{,}000$ steps on the four-source mixture. The second continues for $10{,}000$ steps, and its final checkpoint initializes the sign-projected post-training stage and serves as its frozen teacher.

The final stage starts from this mixed-source checkpoint and trains on Cultura through the sign-projected table; its two loss terms are given in Equation~\ref{eq:starse-loss}.
Its hyperparameters were selected by an exploratory Optuna sweep using the full RuMTEB mean, after which the selected configuration was retrained for a longer run.
This makes the final RuMTEB score selection-biased rather than a held-out estimate.
\subsection{Baselines}
\label{sec:baselines}

We compare StaRSE against two broad groups of models.
The first group consists of selected Sentence Transformer (ST) baselines from the measured quality--latency--size trade-off in Figure~\ref{fig:rumteb-cpu-latency}: ruBERT-tiny-turbo,\footnote{\url{https://huggingface.co/sergeyzh/rubert-tiny-turbo}} ruBERT-mini-FRIDA,\footnote{\url{https://huggingface.co/sergeyzh/rubert-mini-frida}} BERTA,\footnote{\url{https://huggingface.co/sergeyzh/BERTA}} FRIDA,\footnote{\url{https://huggingface.co/ai-forever/FRIDA}} and GigaEmbeddings \citep{kolodin-etal-2025-gigaembeddings}.
We choose these models because they lie on the measured Pareto frontier among the contextual encoders in that comparison.

The second group consists of static embedding baselines.
For Russian pretrained word vectors, we include DeepPavlov fastText wiki+lenta\footnote{\url{https://docs.deeppavlov.ai/en/master/features/pretrained_vectors.html}} and Navec hudlit.\footnote{\url{https://github.com/natasha/navec}}
For multilingual static text embeddings, we include Static similarity\footnote{\url{https://huggingface.co/sentence-transformers/static-similarity-mrl-multilingual-v1}} and Potion multilingual\footnote{\url{https://huggingface.co/minishlab/potion-multilingual-128M}}. Together, these baselines cover Russian pretrained word vectors and public multilingual static text embedders with Russian support.
The static baselines give the closest architectural comparison, while the ST baselines show how StaRSE is positioned relative to strong contextual encoders under the measured quality-cost trade-off.

\subsection{Experiments on Russian Vector Representations}
\label{sec:vector-representation-experiments}

The primary benchmark is RuMTEB (v$1.1$), which contains $23$ Russian tasks. Table~\ref{tab:main-comparison} reports the RuMTEB scores grouped by task type.
The StaRSE row is the sign-coded token-table checkpoint used as the main model in this paper.
It reaches a $51.16$ full RuMTEB task mean, $7.98$ points above the strongest included static baseline, while keeping the static lookup--pool--normalize inference path and storing token directions as signs plus scalar norms.
StaRSE leads the included static baselines in every task family except pair classification, where the static models are effectively tied. This family is represented by TERRa, an entailment task whose sentence-pair relations are a poor fit for context-free mean-pooled token tables.
Contextual ST models remain substantially stronger on retrieval-heavy tasks, as expected from their contextual encoders.
The final block of Table~\ref{tab:main-comparison} reports lexical ranking models on the two reranking and three retrieval tasks; these methods do not produce general-purpose sentence vectors, so no overall RuMTEB mean or score for the remaining task types is defined.

\input{tables/main_comparison}

\paragraph{Lexical ranking protocol.}
\label{sec:lexical-baselines}

We compare TF--IDF \citep{sparck-jones-1972-specificity}, BM25 \citep{robertson-zaragoza-2009-bm25}, PL2 \citep{amati-van-rijsbergen-2002-dfr}, and a Dirichlet-smoothed query-likelihood model \citep{zhai-lafferty-2004-smoothing} with StaRSE's ruBERT WordPiece tokenizer and no stemming or lemmatization. The RuMTEB runs share one analyzer and retain stopwords. Retrieval searches the full corpus, whereas reranking uses each task's supplied candidate pool. We use fixed settings on RuMTEB. BM25 is the strongest lexical baseline in both RuMTEB groups.

We also include an auxiliary long-document diagnostic.
Static encoders do not run self-attention, so long documents can be processed by streaming token lookups into the same mean-pooling accumulator.
We use the Russian subset of MLDR\footnote{\url{https://huggingface.co/datasets/Shitao/MLDR}}, a multilingual retrieval benchmark with especially long documents, and evaluate on $10{,}000$ Russian corpus documents and $200$ test queries.
Table~\ref{tab:mldr-long-context} compares retrieval with $8{,}192$ retained document tokens against static and contextual references: StaRSE reaches $51.66$ nDCG@$10$, $17.35$ points above the next included static baseline and $0.03$ below USER2-small.
The contextual block reproduces the MLDR-RU values reported in the pinned USER2-small model card, while the static rows are evaluated locally on the fixed pool.\footnote{\url{https://huggingface.co/deepvk/USER2-small/blob/23f65b34cf7632032061f5cc66c14714e6d4cee4/README.md}}

\input{tables/mldr_long_context}
\FloatBarrier

Table~\ref{tab:mldr-length-scaling} varies the retained document length for WordPiece BM25, dense StaRSE, z-score fusion of BM25 with StaRSE, and USER2-small as a compact long-context ST baseline. The shorter-length fusion rows use one fixed weight. At $8{,}192$ tokens, we instead report the strongest measured norm-weighted fusion; its variant and weight were selected on the same test pool, so the comparison is descriptive rather than held-out. StaRSE rises from $32.11$ at $128$ tokens to $52.01$ at $4{,}096$ tokens, then scores $51.66$ at $8{,}192$ tokens, within $0.03$ points of USER2-small. BM25 reaches $69.97$ nDCG@$10$, while the exploratory StaRSE+BM25 fusion reaches $73.07$, an observed $3.10$-point difference.

\input{tables/mldr_length_scaling}
\FloatBarrier

Table~\ref{tab:encoding-efficiency} compares encoder footprint and mean end-to-end single-sentence latency for a fixed $95$-character Cultura RU Edu sentence. Timings include tokenization, pooling, and normalization. CPU measurements use an AMD Ryzen 9 7950X, and GPU measurements use one NVIDIA GeForce RTX 3090 Ti. Static-model GPU latency is omitted because no protocol-matched optimized run is available.

\input{tables/encoding_efficiency}

%% file: tables/main_comparison.tex
\begin{table}[!htbp]
\centering
\caption{RuMTEB (v$1.1$) comparison across the $23$ Russian tasks. Columns report the benchmark mean and task-type means on the $0$--$100$ scale; bold marks the best row within each model block. Lexical rankers are defined only for reranking and retrieval, so their other cells are not applicable.}
\label{tab:main-comparison}
\scriptsize
\setlength{\tabcolsep}{3.0pt}
\begin{tabular}{lcccccccc}
\toprule
Model \emph{(dims)} & Mean & Class. & Cluster. & Multilabel & PairClass. & Rerank & Retrieval & STS \\
\midrule
\multicolumn{9}{c}{Sentence Transformers (STs)} \\
\midrule
ruBERT-tiny-turbo \emph{(312)} & 53.76 & 55.08 & 49.62 & 32.05 & 56.27 & 54.94 & 48.82 & 71.73 \\
ruBERT-mini-FRIDA \emph{(312)} & 63.22 & 64.28 & 59.71 & 47.19 & 60.36 & 65.62 & 64.70 & 72.14 \\
BERTA \emph{(768)} & 69.39 & 71.22 & 64.78 & 56.46 & 65.74 & 69.72 & 73.79 & 73.72 \\
FRIDA \emph{(1536)} & 70.95 & 73.71 & \textbf{67.49} & 52.23 & 66.45 & 71.50 & 76.38 & 74.35 \\
GigaEmbeddings \emph{(2048)} & \textbf{74.16} & \textbf{76.66} & 67.48 & \textbf{57.07} & \textbf{79.57} & \textbf{74.02} & \textbf{81.41} & \textbf{75.75} \\
\midrule
\multicolumn{9}{c}{Static Embedding Models} \\
\midrule
DeepPavlov fastText wiki+lenta \emph{(300)} & 42.85 & 51.87 & 42.52 & 27.31 & 52.46 & 30.53 & 18.78 & 55.53 \\
Navec hudlit \emph{(300)} & 36.94 & 46.60 & 37.31 & 26.82 & \textbf{52.57} & 24.44 & 7.66 & 46.73 \\
Static similarity \emph{(1024)} & 42.28 & 46.12 & 34.72 & 28.09 & 51.59 & 38.00 & 28.12 & 61.72 \\
Potion multilingual \emph{(256)} & 43.18 & 50.12 & 41.39 & 28.80 & 50.33 & 33.47 & 22.03 & 59.00 \\
StaRSE \emph{(512)} & \textbf{51.16} & \textbf{56.81} & \textbf{51.80} & \textbf{35.01} & 52.50 & \textbf{41.88} & \textbf{39.09} & \textbf{62.17} \\
\midrule
\multicolumn{9}{c}{Lexical Ranking Models} \\
\midrule
TF--IDF & \textemdash & \textemdash & \textemdash & \textemdash & \textemdash & 39.55 & 43.70 & \textemdash \\
BM25 & \textemdash & \textemdash & \textemdash & \textemdash & \textemdash & \textbf{42.76} & \textbf{46.45} & \textemdash \\
PL2 & \textemdash & \textemdash & \textemdash & \textemdash & \textemdash & 38.34 & 41.91 & \textemdash \\
DirichletLM & \textemdash & \textemdash & \textemdash & \textemdash & \textemdash & 41.06 & 42.44 & \textemdash \\
\bottomrule
\end{tabular}
\end{table}

%% file: tables/mldr_long_context.tex
\begin{table}[!htbp]
\centering
\caption{MLDR-RU long-document retrieval on the test split. Scores are nDCG@$10$ percentages; bold marks the best row within each block.}
\label{tab:mldr-long-context}
\scriptsize
\setlength{\tabcolsep}{4.2pt}
\begin{tabular}{lr}
\toprule
Model \emph{(dims)} & nDCG@$10$ \\
\midrule
\multicolumn{2}{c}{Sentence Transformers (STs)} \\
\midrule
USER2-small \emph{(384)} & 51.69 \\
USER2-base \emph{(768)} & 54.17 \\
USER-bge-m3 \emph{(1024)} & \textbf{58.53} \\
KaLM-v1.5 \emph{(896)} & 53.75 \\
jina-embeddings-v3 \emph{(1024)} & 49.67 \\
E5-mistral-7b \emph{(4096)} & 52.40 \\
\midrule
\multicolumn{2}{c}{Static Embedding Models} \\
\midrule
DeepPavlov fastText wiki+lenta \emph{(300)} & 26.04 \\
Navec hudlit \emph{(300)} & 12.67 \\
Static similarity \emph{(1024)} & 34.31 \\
Potion multilingual \emph{(256)} & 32.95 \\
\textbf{StaRSE \emph{(512)}} & \textbf{51.66} \\
\bottomrule
\end{tabular}
\end{table}

%% file: tables/mldr_length_scaling.tex
\begin{table}[!htbp]
\centering
\caption{MLDR-RU length scaling on $10{,}000$ Russian documents and $200$ test queries. Scores are nDCG@$10$ percentages. Rows show how retrieval quality changes as more document tokens are retained; bold marks the best model at each length.}
\label{tab:mldr-length-scaling}
\scriptsize
\setlength{\tabcolsep}{6pt}
\begin{tabular}{rrrrr}
\toprule
Max doc tokens & BM25 & StaRSE & StaRSE+BM25 & USER2-small \\
\midrule
128  & 28.34 & 32.11 & 30.84 & \textbf{42.31} \\
512  & 30.09 & 33.63 & 34.17 & \textbf{46.97} \\
1024 & 32.47 & 36.88 & 36.05 & \textbf{48.22} \\
2048 & 34.07 & 40.90 & 38.38 & \textbf{48.67} \\
4096 & 68.12 & 52.01 & \textbf{69.93} & 53.70 \\
8192 & 69.97 & 51.66 & \textbf{73.07} & 51.69 \\
\bottomrule
\end{tabular}
\end{table}

%% file: tables/encoding_efficiency.tex
\begin{table}[!htbp]
\centering
\caption{RuMTEB score, encoder footprint, and mean end-to-end single-sentence latency on a fixed Cultura RU Edu input. Contextual footprints are estimated from 16-bit parameter counts; static footprints are measured packages. Static-model GPU latency is omitted because no protocol-matched optimized run is available. Bold marks the best reported value within each block and column.}
\label{tab:encoding-efficiency}
\scriptsize
\setlength{\tabcolsep}{4.5pt}
\begin{tabular}{lrcrr}
\toprule
Model \emph{(dims)} & Size (MiB) & RuMTEB & CPU (ms) & GPU (ms) \\
\midrule
\multicolumn{5}{c}{Sentence Transformers (STs)} \\
\midrule
ruBERT-tiny-turbo \emph{(312)} & \textbf{55.7} & 53.76 & \textbf{1.43} & \textbf{1.68} \\
ruBERT-mini-FRIDA \emph{(312)} & 61.5 & 63.22 & 2.97 & 3.01 \\
BERTA \emph{(768)} & 245 & 69.39 & 19.36 & 3.80 \\
FRIDA \emph{(1536)} & 1571 & 70.95 & 110.87 & 9.61 \\
GigaEmbeddings \emph{(2048)} & 6580 & \textbf{74.16} & 773.61 & 43.26 \\
\midrule
\multicolumn{5}{c}{Static Embedding Models} \\
\midrule
DeepPavlov fastText wiki+lenta \emph{(300)} & 1853 & 42.85 & 0.05 & -- \\
Navec hudlit \emph{(300)} & 589 & 36.94 & \textbf{0.04} & -- \\
Static similarity \emph{(1024)} & 414 & 42.28 & 0.17 & -- \\
Potion multilingual \emph{(256)} & 507 & 43.18 & 0.07 & -- \\
StaRSE \emph{(512)} & \textbf{11.3} & \textbf{51.16} & 0.08 & -- \\
\bottomrule
\end{tabular}
\end{table}

%% file: sections/05_analysis.tex
\section{Analysis}

\subsection{Vocabulary, Initialization, and PCA Selection}
\label{sec:model-selection}

We first choose the tokenizer vocabulary, because it determines the set of trainable vectors in a static embedding model. We screen Russian-focused tokenizers with two lightweight diagnostics: the number of retained analyzer tokens per lexical word and BM25 nDCG@$10$ when each tokenizer is used as a tokenizer-only analyzer \citep{robertson-zaragoza-2009-bm25} on $12$ RusBEIR retrieval tasks \citep{kovalev-etal-2025-rusbeir}. The screen focuses on Russian tokenizers; token-count statistics use a sample of $250{,}000$ lexical words extracted from RusBEIR text fields, and tokenizers with identical sample tokenization are represented by one family. The ruBERT-base tokenizer \citep{zmitrovich-etal-2023-family} gives the best lexical proxy score while keeping segmentation reasonably compact, so StaRSE uses its $120{,}138$-token vocabulary. Table~\ref{tab:vocabulary-selection} reports the full diagnostics.

The valid-token count is measured after analyzer filtering, including removal of special and punctuation tokens, and is not an unfiltered tokenizer fertility measure. The word-frequency baseline emits one token for a word present in its vocabulary and none for an out-of-vocabulary word. Its $0.813$ valid tokens per word therefore reflects $81.35\%$ word coverage ($203{,}366$ of $250{,}000$ sampled words), with the remaining $18.65\%$ omitted. It emits exactly one token per covered word, so the sub-unit value is not a segmentation advantage over the subword tokenizers.

\input{tables/vocabulary_selection}

After fixing the vocabulary, we screen matched $256$-dimensional initialization choices on the validation split of the Russian STS benchmark.\footnote{\url{https://huggingface.co/datasets/ai-forever/ru-stsbenchmark-sts}} Each candidate is converted into a static embedding table and evaluated by cosine Spearman. The highest development score comes from ruBERT-base input embeddings followed by token-level PCA, ABTT, and SIF-Zipf weighting. It scores above the matched sentence-level PCA variant \citep{wada-etal-2025-static} and contextual-token alternatives in this screen, so we transfer the same input-embedding/token-level-PCA recipe to the final $512$-dimensional StaRSE model. Table~\ref{tab:init-selection} gives the detailed results.

\input{tables/initialization_selection}
\FloatBarrier

\subsection{Signal in Adjacent Sentence Pairs}
\label{sec:adjacent-pair-signal}

The method relies on adjacent sentences as weak positives, so we first check whether this signal is stronger than random document-level co-occurrence. We score three Cultura pair distributions with USER2-small used only as an external diagnostic encoder: random cross-document pairs, unfiltered adjacent sentence pairs, and adjacent pairs retained by the lemma-overlap filter from Section~\ref{sec:method-mining}. This encoder is not used to train StaRSE.

We sample $30{,}000$ pairs from each distribution. The median USER2-small cosine similarity is $0.573$ for random cross-document pairs, $0.726$ for unfiltered adjacent pairs, and $0.820$ after the lemma-overlap filter. The filter is not a semantic objective: it favors lexical overlap, while the model itself is trained with broad contrastive batches.

\subsection{Ablations}
\label{sec:ablations}

We run two groups of ablations. The first tests data construction and distillation baselines, while the second isolates the final training stage that uses the sign-projected token table.

\paragraph{Data and distillation ablations.}
We first test two design choices under a simpler setup than the final StaRSE training. The self-pair, raw-adjacent, and lemma-overlap-mix contrastive runs process approximately $10$M pairs each. Table~\ref{tab:data-ablation} is therefore a set of single-run diagnostics, not a replacement for the main results.

\input{tables/data_ablation}

The first block reports two static-student distillation baselines at $42.0$ and $44.2$ RuMTEB, alongside the higher non-self contrastive scores of $48.2$ and $49.9$ in the second block. For the distillation baselines, we follow the SWE-style algorithm of \citet{wada-etal-2025-static}: a static student is trained to match the pairwise similarity distribution induced by an ST teacher. We adapt it to the Russian teachers by using each teacher's native tokenizer and vocabulary rather than constructing a new word-level vocabulary. We use STs as teachers because they already define a sentence-level semantic space; ruBERT-base is not used as a distillation teacher because it is a masked language model and serves here only as the source of the vocabulary and input-embedding initialization. Because the two blocks use different training budgets and no seed replication, we treat them as complementary diagnostics rather than a controlled ranking of objectives.

The second block examines positive-pair construction. Self-pairs score lowest, while raw adjacent Cultura and the lemma-overlap mix ($\geq0.20$) reach $48.2$ and $49.9$ RuMTEB, respectively; their RuSTS development scores are $76.1$ and $77.6$. The lemma-overlap mix has the highest observed scores in this block. These single-run diagnostics do not isolate the effect of the overlap filter from the source mixture.

\paragraph{Sign-projected table and inference ablations.}
The final StaRSE checkpoint is sign-coded at the token-table level, so the key question is whether this compact table should be trained directly or obtained by post-hoc signing.
Table~\ref{tab:sign-projection-ablations} collects the relevant checks.
In the same $512$-dimensional architecture, signing the table immediately after contrastive fine-tuning gives $49.33$ RuMTEB.
Continuing ordinary contrastive fine-tuning before signing scores $50.09$, while quantization-aware post-training through the sign-projected table reaches $51.16$.
The same table reports three inference-time negative controls. Dropping token norms is much more damaging: replacing norm-scaled sign directions with unweighted signs costs $8.42$ points.
Keeping token norms but signing the pooled sentence vector costs $3.84$ points.
The compact StaRSE representation therefore stores token directions as signs plus scalar norms, then returns a real-valued $\ell_2$-normalized sentence vector after mean pooling.

\input{tables/sign_projection_ablations}

%% file: tables/vocabulary_selection.tex
\begin{table}[t]
\centering
\caption{Tokenizer diagnostics used for vocabulary selection.
BM25 is used as a tokenizer-only retrieval proxy over $12$ RusBEIR tasks.
BM25 nDCG@$10$ is reported on the $0$--$100$ scale.
Valid tok/word counts retained analyzer tokens per sampled lexical word and is affected by filtering and out-of-vocabulary coverage.}

\label{tab:vocabulary-selection}
\scriptsize
\setlength{\tabcolsep}{5pt}
\begin{tabular}{lccc}
\toprule
Tokenizer & Vocab & Valid tok/word & BM25 nDCG@$10$ \\
\midrule
\texttt{ai-forever/ruBert-base} & 120,138 & 1.342 & \textbf{52.46} \\
\texttt{deepvk/USER-bge-m3} & 46,166 & 1.696 & 51.02 \\
\texttt{cointegrated/rubert-tiny2} & 83,828 & 1.303 & 50.81 \\
\texttt{DeepPavlov/rubert-base-cased-conversational} & 100,792 & 1.280 & 50.57 \\
\texttt{cointegrated/rubert-tiny} & 29,564 & 2.171 & 50.27 \\
\texttt{deepvk/USER2-small} & 50,368 & 2.153 & 49.88 \\
\texttt{ai-forever/FRIDA} & 93,651 & 2.130 & 49.79 \\
\texttt{DeepPavlov/rubert-base-cased} & 119,547 & 1.264 & 49.74 \\
\texttt{ai-forever/ru-en-RoSBERTa} & 98,505 & 2.135 & 49.72 \\
\texttt{ai-sage/Giga-Embeddings-instruct} & 128,256 & 2.213 & 49.67 \\
\texttt{ai-forever/ruElectra-large} & 120,138 & 1.329 & 49.64 \\
\texttt{ai-forever/ruRoberta-large} & 50,257 & 2.152 & 49.43 \\
Top 150k Russian word-frequency vocabulary & 150k & 0.813 & 42.18 \\
\bottomrule
\end{tabular}
\end{table}

%% file: tables/initialization_selection.tex
\begin{table}[t]
\centering
\caption{Matched initialization screening on the Russian STS Benchmark validation split. All rows use the ruBERT-base vocabulary with the same post-processing. Scores are cosine Spearman; bold marks the selected initialization.}
\label{tab:init-selection}
\small
\begin{tabular}{llc}
\toprule
Initialization source & PCA & STS dev \\
\midrule
Input embeddings & token-level & \textbf{0.76810} \\
Input embeddings & sentence-level & 0.73663 \\
Averaged contextual embeddings & sentence-level & 0.73349 \\
Per-token last-hidden state (Model2Vec) & token-level & 0.64753 \\
\bottomrule
\end{tabular}
\end{table}

%% file: tables/data_ablation.tex
\begin{table}[t]
\centering
\caption{Data-construction and distillation ablations. Scores are percentages; bold marks the best result in each column.}
\label{tab:data-ablation}
\small
\setlength{\tabcolsep}{8pt}
\begin{tabular}{lcc}
\toprule
Variant & RuSTS dev & RuMTEB \\
\midrule
\multicolumn{3}{l}{\textit{Distillation baselines}} \\
\midrule
BERTA-distilled static student & 74.7 & 42.0 \\
USER2-small-distilled static student & 69.2 & 44.2 \\
\midrule
\multicolumn{3}{l}{\textit{Positive-pair construction}} \\
\midrule
Self-pairs & 71.1 & 37.2 \\
Raw adjacent Cultura & 76.1 & 48.2 \\
Lemma-overlap mix ($\geq0.20$) & \textbf{77.6} & \textbf{49.9} \\
\bottomrule
\end{tabular}
\end{table}

%% file: tables/sign_projection_ablations.tex
\begin{table}[t]
\centering
\caption{Ablations for the sign-coded StaRSE token table. Scores are full-RuMTEB means; deltas are relative to the first row in each block.}
\label{tab:sign-projection-ablations}
\small
\setlength{\tabcolsep}{5pt}
\begin{tabular}{p{0.70\linewidth}rr}
\toprule
Setting & RuMTEB & $\Delta$ \\
\midrule
\multicolumn{3}{l}{\textit{Training route to a sign-coded token table}} \\
Post-hoc signing after contrastive fine-tuning; no post-training & 49.33 & -- \\
Contrastive continuation, then post-hoc signing & 50.09 & +0.76 \\
Optimized sign-projected post-training, final StaRSE recipe & 51.16 & +1.83 \\
\midrule
\multicolumn{3}{l}{\textit{Inference-time negative controls on the final checkpoint}} \\
StaRSE inference: token norms kept; real-valued pooled sentence vector & 51.16 & -- \\
Drop token norms: use unweighted $\operatorname{sign}(u_i)/\sqrt d$ rows & 42.74 & $-8.42$ \\
Drop token norms and sign the pooled sentence vector & 39.06 & $-12.10$ \\
Keep token norms but sign the pooled sentence vector & 47.33 & $-3.84$ \\
\bottomrule
\end{tabular}
\end{table}

%% file: sections/06_conclusion.tex
\section{Conclusion}

We presented StaRSE, a lightweight Russian static embedding model based on a compact sign-coded token table, mean pooling, and contrastive training on Russian sentence pairs. The final checkpoint reaches a $51.16$ RuMTEB task mean while storing token directions as binary signs plus scalar norms in a measured $11.3$~MiB package. It scores above the included static baselines on the RuMTEB retrieval tasks, and its quantization-aware post-training gives the strongest RuMTEB mean among the evaluated routes to a sign-coded token table. On the long-document MLDR-RU diagnostic, StaRSE reaches $51.66$ nDCG@$10$, close to USER2-small at $51.69$, while retaining a streaming lookup-and-pooling encoder. Future work should add held-out hyperparameter selection and replicated runs, strengthen retrieval-specific supervision, measure end-to-end retrieval-system efficiency, and identify the domains and query types for which lexical and StaRSE signals are most complementary. It should also explore whether strict one-bit-per-dimension sentence representations can reduce the remaining gap to real-valued pooled sentence vectors.